\documentclass[conference]{IEEEtran}
\IEEEoverridecommandlockouts

\usepackage{amsmath,amssymb,amsfonts}
\usepackage{float}
\usepackage{multirow}
\usepackage{booktabs}
\usepackage{algorithmic}
\usepackage{graphicx}
\usepackage{textcomp}
\usepackage{xcolor}
\def\BibTeX{{\rm B\kern-.05em{\sc i\kern-.025em b}\kern-.08em
    T\kern-.1667em\lower.7ex\hbox{E}\kern-.125emX}}

\usepackage[
backend=biber,
sorting=none
]{biblatex}
\begin{document}

\title{GCS-M3VLT: Guided Context Self-Attention based Multi-modal Medical Vision Language Transformer for Retinal Image Captioning}

\author{
Teja Krishna Cherukuri\textsuperscript{$\dag\ast$} \quad 
Nagur Shareef Shaik\textsuperscript{$\dag\ast$} \quad 
Jyostna Devi Bodapati\textsuperscript{$\ddag$} \quad
Dong Hye Ye\textsuperscript{$\dag$} \\\\
\textsuperscript{$\dag$}Department of Computer Science, Georgia State University, Atlanta, GA, USA \\
\textsuperscript{$\ddag$}Vignan's Foundation for Science, Technology \& Reserach University, Guntur, Andhra Pradesh, India \\
\thanks{\textsuperscript{$\ast$}Authors contributed equally}\thanks{Corresponding Author: dongye@gsu.edu}
}


\maketitle

\begin{abstract}
Retinal image analysis is crucial for diagnosing and treating eye diseases, yet generating accurate medical reports from images remains challenging due to variability in image quality and pathology, especially with limited labeled data. Previous Transformer-based models struggled to integrate visual and textual information under limited supervision. In response, we propose a novel vision-language model for retinal image captioning that combines visual and textual features through a guided context self-attention mechanism. This approach captures both intricate details and the global clinical context, even in data-scarce scenarios. Extensive experiments on the DeepEyeNet dataset demonstrate a 0.023 BLEU@4 improvement, along with significant qualitative advancements, highlighting the effectiveness of our model in generating comprehensive medical captions.
\end{abstract}


\section{Introduction}

The rising incidence of retinal diseases like Diabetic Retinopathy (DR) and Diabetic Macular Edema (DME) poses a global health challenge, with DR alone affecting nearly a one-third of diabetic individuals and leading to vision-threatening complications in about 10\% of cases \cite{ciulla2003diabetic, shaik2022hinge, sommer2014challenges}. Early detection is crucial, yet traditional diagnosis methods using Color Fundus Photography and Optical Coherence Tomography are resource-intensive, relying heavily on ophthalmologists \cite{pizzarello2004vision}. Automating medical report generation from retinal images offers a potential solution, but remains difficult due to the complexity of retinal pathologies, image variability, and limited annotated data \cite{huang2021deepopht}. Current approaches often borrow from general image captioning models \cite{xu2015show, you2016image}, yet struggle to meet the high accuracy and interpretability demands in medical applications \cite{zhang2024sam, huang2023kiut}.

Models like DeepOpht \cite{huang2021deepopht} and Deep Context-Encoding \cite{huang2021deep} made early progress in retinal image captioning but struggled due to their simplistic architectures, limiting their ability to capture complex visual-clinical interactions. The integration of attention mechanisms in models like Non-local Attention \cite{huang2022non}, Contextualized GPT \cite{huang2021contextualized}, and Expert Transformer \cite{wu2023expert} enhanced coherence and knowledge integration but still struggled with multi-modal fusion in complex medical scenarios. More recent advancements, such as the Gated Contextual Transformer \cite{shaik2024gated} and M3 Transformer \cite{shaik2024m3t}, have improved context-based medical image captioning, particularly in multi-modal fusion \cite{cao2023mmtn, wu2024mm}.

Parallel developments in Vision-Language Models (VLMs) have also contributed to this field. The ME Transformer \cite{wang2023metransformer} introduced learnable expert tokens for better modality adaptation, yet its high computational cost limits practical deployment. Similarly, VisionGPT \cite{kelly2024visiongpt} and LlaVA \cite{liu2024visual} offer robust vision-language integration but require significant computational resources. To overcome these limitations, we propose the Guided Context Self-Attention based Multi-modal Medical Vision Language Transformer (GCS-M3VLT), aimed at addressing modality integration and computational efficiency in retinal image report generation.

\begin{figure*}[!t]
    \centerline{\includegraphics[width=0.9\textwidth]{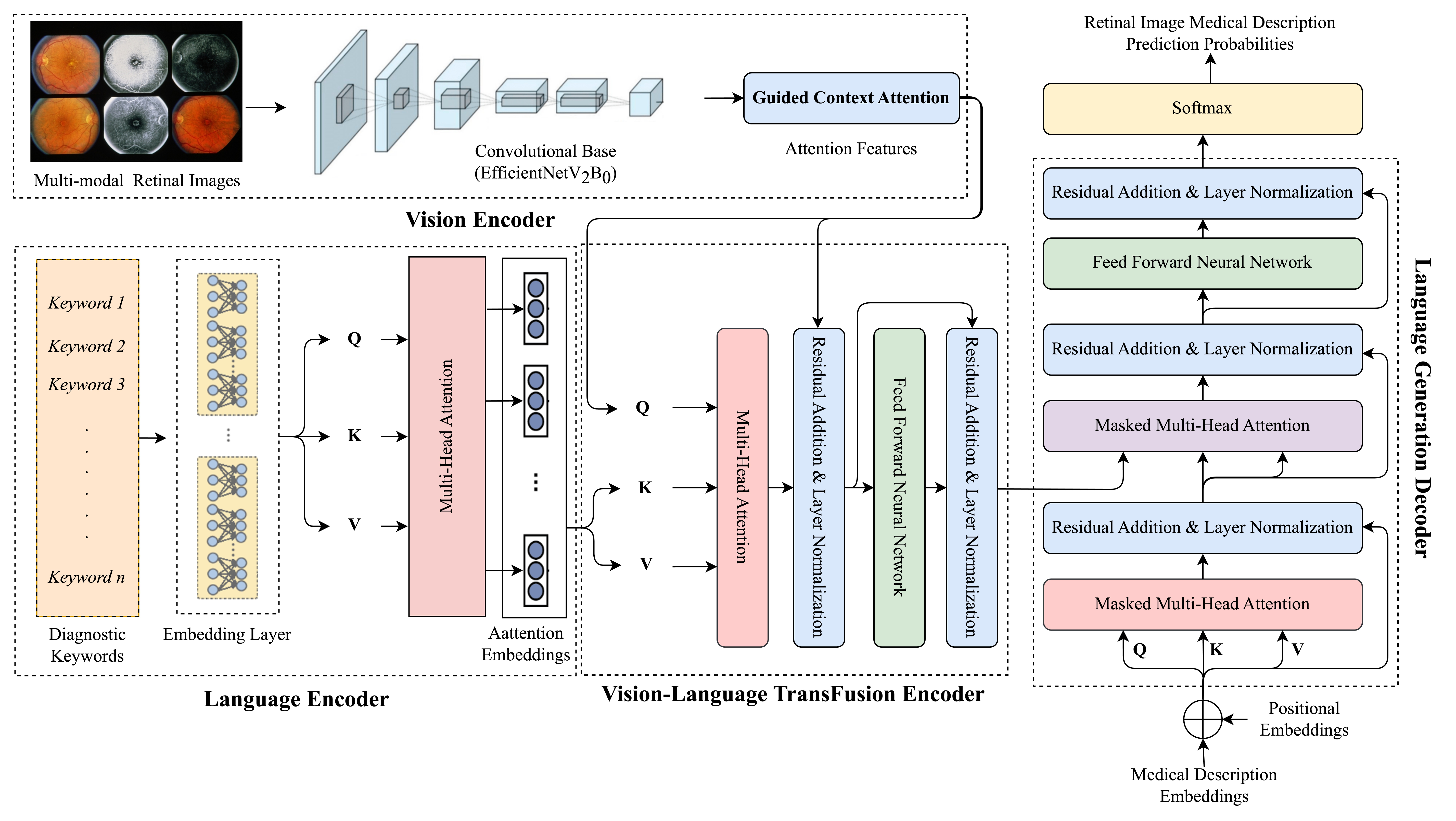}\vspace{-10pt}}
    \caption{Architecture of the proposed Guided Context Self-Attention-based Multi-modal Medical Vision Language Transformer (GCS-M3VLT); \textbf{Vision Encoder} -- Learns attention-based representations from retinal images to capture visual features crucial for diagnosis; \textbf{Language Encoder} -- Learns self-attention-based clinical-context embeddings from diagnostic keywords, enabling the model to understand the semantic context of medical terms; \textbf{Vision-Language TransFusion Encoder} -- Integrates visual attention features and clinical-context embeddings, leveraging both visual and semantic information to provide comprehensive understanding; \textbf{Language Generation Decoder} -- Generates coherent and meaningful medical descriptions by attending to relevant visual and semantic cues, ensuring contextually appropriate and diagnostically relevant outputs.}
    \vspace{-10pt}
    \label{fig:GCS-M3VLT}
\end{figure*}

\section{Methodology}

This research aims to derive clinical context from multi-modal retinal images and diagnostic keywords to generate accurate medical captions. We introduce the Guided Context Self-Attention based Multi-modal Medical Vision Language Transformer (GCS-M3VLT), which fuses visual and textual data for comprehensive medical descriptions. Figure \ref{fig:GCS-M3VLT} illustrates the model architecture, with technical details provided in the following subsections.

\vspace{-5pt}
\subsection{Vision Encoder}

The Vision Encoder converts pre-processed retinal images into visual features using a Convolutional base and a Guided Context Attention mechanism. The Convolutional base, employing EfficientNetV2B0 \cite{tan2021efficientnetv2}, extracts initial spatial features from images. Given a retinal scan image $X_R$ of dimensions $(356 \times 356 \times 3)$, Convolutional base produces visual features $F_R$ of dimensions $(12 \times 12 \times 1280)$.

\subsubsection{Guided Context Attention}

The Guided Context Attention (GCA) block enhances retinal image analysis by integrating spatial and channel contexts. This approach is essential for capturing lesion-specific details that traditional attention mechanisms may miss \cite{cherukuri2024guided}. We start with spatial features from the convolutional base, formulated as Context Query $(Q_c)$, Context Key $(K_c)$, and Context Value $(V_c)$. The GCA block refines these features through two main stages: spatial and channel context formulation.

\paragraph{Spatial Context} Spatial context is computed using global attention pooling:
\vspace{-8pt}
\begin{equation}
    F_{s} = \sum\limits_{j=1}^d \frac{e^{W_r f_{r,j}}}{\sum\limits_{m=1}^d e^{W_r f_{r,m}}} \qquad \forall f_{r,j} \in F_R
    \label{eq:fs}
\end{equation}
where $F_s$ represents spatial context features, and $W_r$ are the point-wise convolution parameters. This operation normalizes the spatial features, emphasizing essential global information.

\paragraph{Channel Context} Channel context is incorporated as follows: 
\vspace{-10pt}
\begin{equation}
    F_{c} = F_R \oplus \sum\limits_{i=1}^d W_{2}\left(\text{LN}\left(\Gamma\left(\sum\limits_{j=1}^{k} W_{1}f_{s_j}\right)\right)\right)_i
    \forall f_{s_j} \in F_{s}
    \label{eq:fc}
\end{equation}
Here, $F_c$ combines spatial context with channel-wise features, where $W_1$ and $W_2$ are point-wise convolutions, $\Gamma$ denotes ReLU, and LN stands for Layer Normalization.

\paragraph{Attention Coefficients} The final attention-weighted features are obtained through: \begin{equation} F_{gca} = \sigma(W_\psi \cdot \Gamma(W_{qc} Q_c + W_{kc} K_c + b_{qk}) + b_{\psi}) \cdot V_c \label{eq
} \end{equation} where $F_{gca}$ are the refined visual features, with $W_{qc}$, $W_{kc}$, $b_{qk}$, $W_\psi$, and $b_{\psi}$ as the gating parameters. This formula combines $Q_c$ and $K_c$ to compute attention coefficients, which are applied to $V_c$ to focus on relevant features. 
\vspace{-10pt}
\begin{figure}[H]
    \centerline{\includegraphics[width=0.5\textwidth]{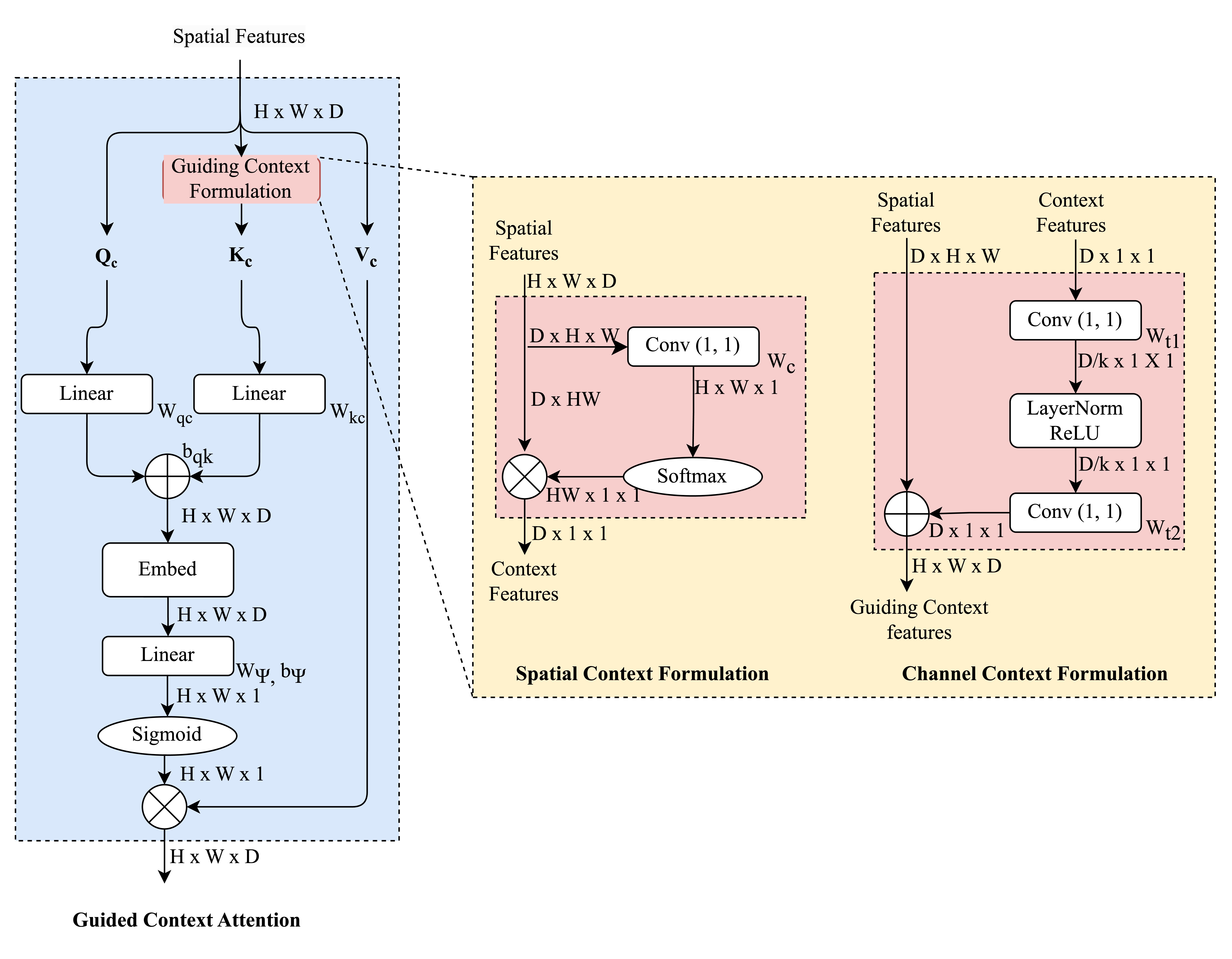} \vspace{-10pt}}
    \caption{Architecture of Guided Context Attention that utilizes context features to compute lesion contextual attention representations; \textbf{Spatial Context Formulation} -- selectively focuses on relevant spatial features in the initial representations and computes global context information; \textbf{Channel Context Formulation} -- processes the computed context information \& capture channel-wise correlations;}
    \vspace{-10pt}
    \label{fig:GCA}
\end{figure}

\subsection{Language Encoder}

The language encoder transforms processed diagnostic keywords into attention embeddings. It employs the Embedding layer to convert the processed diagnostic keyword sequence into embeddings, denoted as $KE$, encapsulating individual keyword embeddings $ke_1, ke_2, ..., ke_n$. The encoder captures the semantic meaning of input keywords, but the generated embeddings initially lack explicit relationships between keywords, making them context-free. To overcome this limitation, the embeddings undergo further processing by a Multi-Head Attention. This layer employs a scaled-dot product attention mechanism, wherein three matrices are computed: Queries $(Q = KE)$, Keys $(K = KE)$, and Values $(V = KE)$. These matrices are subsequently processed to obtain context-enriched keyword embeddings $(KE_c)$ that encapsulate meaningful relationships between the keywords . 
\begin{equation}
    \text{H}_i = \text{Self-Attention}(KE \cdot W_{q_i}, KE \cdot W_{k_i}, KE \cdot W_{v_i}) \label{eq:head}
\end{equation}
\vspace{-5mm}
\begin{equation}
    \text{Self-Attention}(Q, K, V) = \text{Softmax}\left(\frac{QK^T}{\sqrt{d_k}}\right)V \label{eq:self-att}
\end{equation}
\vspace{-5mm}
\begin{equation}
    K_{att} = \text{MHA}(KE) = \text{Concat}(\text{H}_1, \text{H}_2, ..., \text{H}_h)W_o \label{eq:mha}
\end{equation}
Equations \ref{eq:head} to \ref{eq:mha} show the computation of attention for each head in the MHA mechanism, with \ref{eq:self-att} detailing the Self-Attention scores and \ref{eq:mha} defining the concatenation of attention outputs from all heads. In these operations, $KE$ represents the input embeddings; $W_{q_i}$, $W_{k_i}$, and $W_{v_i}$ are the weight matrices associated with the keyword embeddings for $Q$, $K$, and $V$ in the $i$-th head, respectively; $\sqrt{d_k}$ is the length of $K$; $W_o$ is the output weight matrix and, $h$ denotes the number of heads. The final embeddings are derived from the fusion of the input embedding and context embedding. This fusion is achieved by element-wise addition and normalization, represented by the equation \ref{eq:kefinal} where $\oplus$ denotes the element-wise addition operation.
\begin{equation}
    KE_{final} = \text{LN}\left(KE \oplus \text{MHA}(KE)\right) \label{eq:kefinal}
\end{equation}

\subsection{Vision-Language TransFusion Encoder}

The Vision and Language encoders generate intermediate features within their domains, and the TransFusion Encoder (TFE) attentively fuses keyword context embeddings $(KE_{final})$ with visual representations $(F_{gca})$ for comprehensive clinical understanding. This fusion enables the cross-modality interactions necessary for accurate caption generation. Within the TFE, we implement a Vision-Language Cross-Attention (VLA) mechanism using Multi-Head Attention. The VLA selectively incorporates features from both domains, ensuring that critical clinical context is preserved and emphasized. This mechanism dynamically adjusts the contributions of visual and textual features, assigning higher weights to informative elements while suppressing irrelevant or redundant information, thus enhancing the relevance of generated captions. The VLA mechanism is formalized in Equation \ref{eq:vla}:
\vspace{-4pt}
\begin{equation}
   Z = \text{MHA}(F_{gca}, KE_{final}, KE_{final})
   \label{eq:vla}
\end{equation}
\vspace{-5mm}
\begin{equation}
    F^{'} = \text{LN}(W_r \Gamma(W_h, \text{LN}(Q + Z)) + \text{LN}(Q + Z)) \label{eq:f'}
\end{equation}
Subsequent operations, including Residual Addition, Layer Normalization (LN), and a Feed Forward Neural Network (FFNN) involving ReLU $(\Gamma)$, with parameters $W_h$ and $W_r$, refine the attention representations, resulting in $F^{'}$.
\vspace{-5pt}
\subsection{Language Generation Decoder}
The final component of GCSM3VLT is the Language Generation Decoder, which translates the fused multi-modal features into clinically meaningful textual descriptions. This decoder is based on a Transformer (GPT-2) \cite{vaswani2017attention, radford2019language} that incorporates both the original visual representations and the integrated Vision-Language features generated by the TransFusion Encoder. By leveraging a sequence of multi-head attention layers, the decoder constructs a contextualized representation of the input modalities, which is then used to predict the next word in the sequence until the entire caption is generated.

\section{Experiments}

\vspace{-3pt}

\begin{figure*}[!ht]
    \centerline{\includegraphics[width=\textwidth]{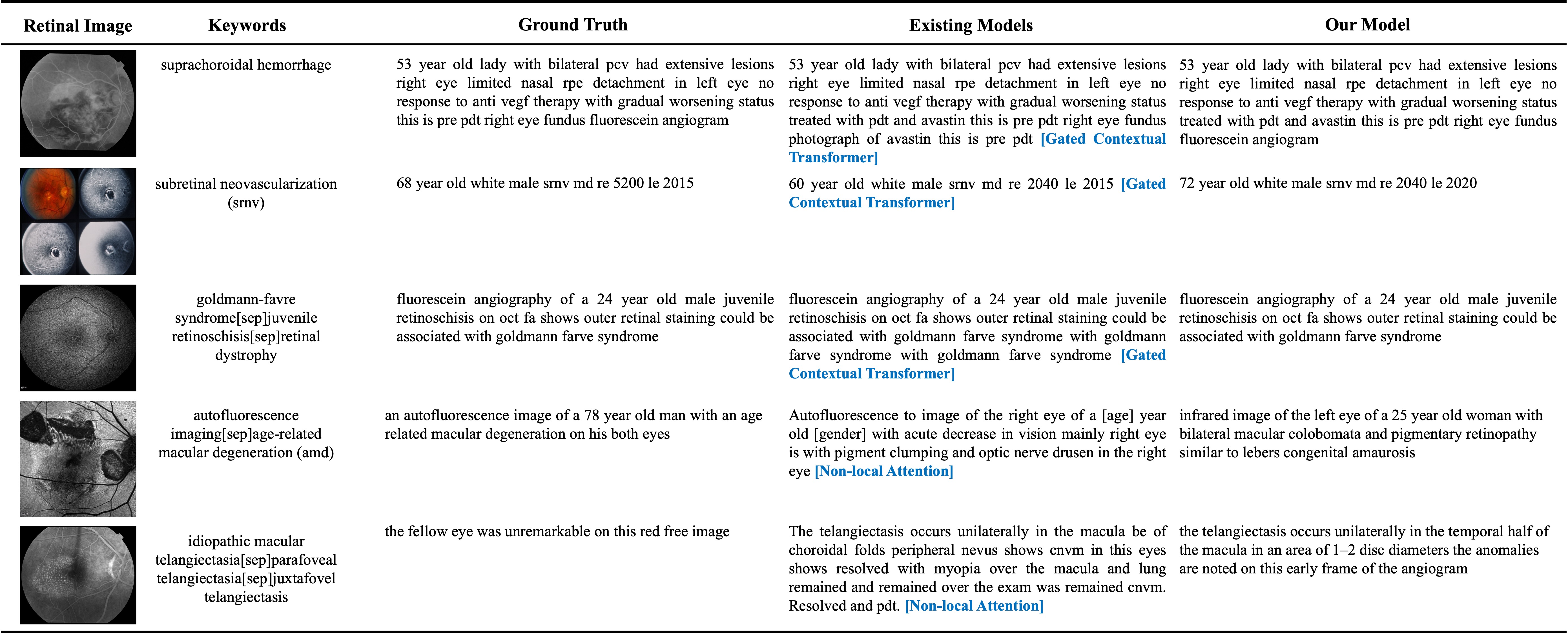} \vspace{-5pt}}
    \caption{Comparison of Actual and Predicted Captions obtained using the proposed GCS-M3VLT with Existing Works \cite{shaik2024gated}\cite{huang2022non} using Retinal Image and Keywords as input. The captions generated by GCS-M3VLT closely resemble the ground truth captions, showcasing the model's ability to accurately describe the clinical context of retinal images. In contrast, captions from existing works exhibit discrepancies and lack coherence compared to ground truth, underscoring the superior performance of the proposed approach. \vspace{-10pt}}
    \label{fig:Captions-comparision}
\end{figure*}

\subsection{Dataset \& Implementation Details} To evaluate GCSM3VLT, we use the DeepEyeNet dataset \cite{huang2021deepopht} comprising 15,710 retinal images across various modalities: Fluorescein Angiography, Fundus Photography, Optical Coherence Tomography, and multi-modality grids. Annotated by expert ophthalmologists with 609 unique diagnostic keywords and clinical descriptions averaging 5-10 words. Covering 265 retinal diseases, from common to rare cases, the dataset is split into training (60\%), validation (20\%), and testing (20\%) subsets, offering a robust foundation for high-precision diagnostic model development. All Images were resized to $(356 \times 356)$ pixels with three channels, and text was cleaned and standardized, with keywords limited to 5-50 words, captions to 50 words, and rare words replaced by $<$UNK$>$. The vocabulary was capped at 5000 tokens, with 1024-dimensional embeddings. The model was trained for 100 epochs (loss: cross\_entropy, optimizer: Adam, batch size: 64, learning rate: 0.0001) on an NVIDIA P100 GPU. Performance was evaluated using BLEU, CIDEr, and ROUGE metrics.

\subsection{Quantitative Evaluation}

Table \ref{tab:comparision_study} highlights the impressive performance of our GCS-M3VLT, which surpasses state-of-the-art Vision-Language Models (VLMs) across all evaluation metrics. Our model achieves the highest BLEU scores, outshining leading models such as DeepOpht, Contextualized Keywords, and VisionGPT. It also excels in CIDEr and ROUGE, showcasing its superior ability to generate precise and contextually relevant medical captions compared to VLMs like Gated Contextual Attention Net and M3 Transformer. Additionally, the GCS-M3VLT's lightweight design, with fewer parameters, contrasts favorably with heavier models like Expert Transformer, LlaVA-Med, and VisionGPT. Its Guided Context Self-Attention mechanism proves to be more effective than Non-local and Gated Contextual Attention, emphasizing its enhanced vision-language integration.

\begin{table}[!ht]
  \centering
  \vspace{-10pt}
  \caption{Comparative Study of Recent Best Models with Proposed Guided Context Self-Attention based Multi-modal Medical Vision Language Transformer trained on DeepEyeNet Dataset}
  \label{tab:comparision_study}
  \setlength{\tabcolsep}{1.5pt} 
  \begin{tabular}{lcccccc}
    \toprule
    \textbf{Model} & \textbf{B@1} & \textbf{B@2} & \textbf{B@3} & \textbf{B@4} & \textbf{CIDEr} & \textbf{ROUGE} \\
    \midrule
    DeepOpth \cite{huang2021deepopht} & 0.184 & 0.114 & 0.068 & 0.032 & 0.361 & 0.232 \\
    Deep Context Encoding \cite{huang2021deep} & 0.219 & 0.134 & 0.074 & 0.035 & 0.398 & 0.252 \\
    Contextualized GPT \cite{huang2021contextualized} & 0.203 & 0.142 & 0.100 & 0.073 & 0.389 & 0.211 \\
    Non-local Attention \cite{huang2022non} & 0.230 & 0.150 & 0.094 & 0.053 & 0.370 & 0.291 \\
    Gated Contextual Transformer \cite{shaik2024gated} & 0.297 & 0.230 & 0.214 & 0.142 & 0.462 & 0.391 \\
    VisionGPT \cite{kelly2024visiongpt} & 0.353 & 0.280 & 0.261 & 0.182 & 0.491 & 0.412 \\
    Expert Transformer \cite{wu2023expert} & 0.382 & 0.291 & 0.237 & 0.186 & 0.472 & 0.413 \\
    LlaVA-Med \cite{li2024llava} & 0.386 & 0.305 & 0.282 & 0.196 & 0.482 & 0.427 \\
    M3 Transformer \cite{shaik2024m3t} & 0.394 & 0.312 & 0.291 & 0.208 & 0.537 & 0.429 \\
    \textbf{GCS-M3VLT (Ours)} & \textbf{0.430} & \textbf{0.345} & \textbf{0.319} & \textbf{0.231} & \textbf{0.559} & \textbf{0.497} \\
    \bottomrule
  \end{tabular}
  \vspace{-10pt}
\end{table}

\subsection{Qualitative Evaluation}

\begin{figure}[!ht]
    \centerline{\includegraphics[width=0.45\textwidth]{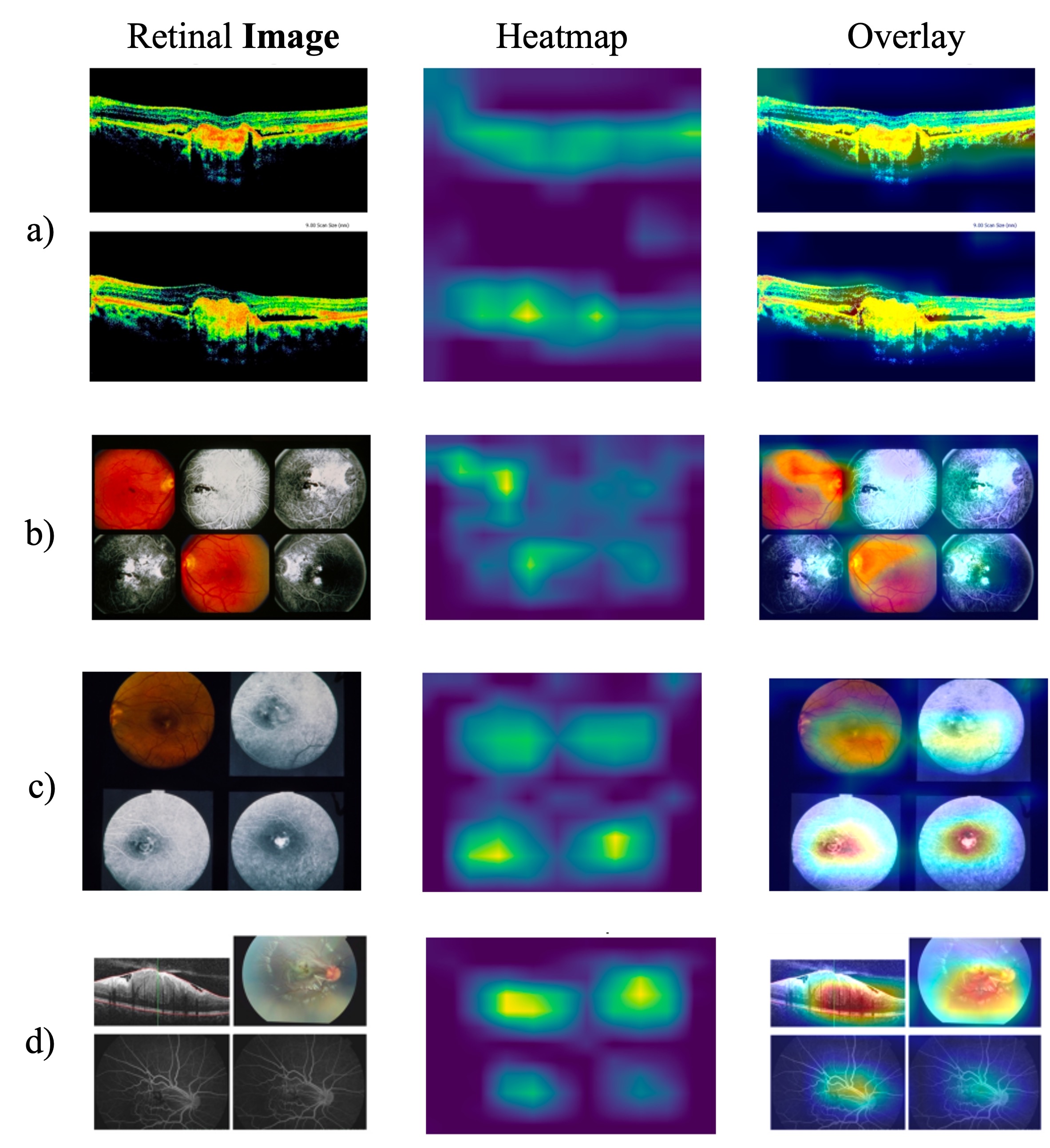} \vspace{-10pt}}
    \caption{Visualizing heatmaps and gradient class activation maps obtained using Vision Encoder with Guided Context Attention features of (a) OCT (b \& c) fundus + AF (d) OCT + fundus + AF retinal images highlighting lesion context information}
    \vspace{-10pt}
    \label{fig:GradCAMPP-MultiModal}
\end{figure}

In Figure \ref{fig:GradCAMPP-MultiModal}, we illustrate the input images, heatmaps, and overlays of attention maps, visualized through GradCAM \cite{selvaraju2017grad}, when the input images consist of a grid of multiple scans with the same or different modalities. For instance, we present scenarios where the input includes OCT scans, fundus images, and AF images, either individually or in combination. Our model robustly captures clinical context and lesion-specific information, even when the input comprises a grid of multiple images. 

Figure \ref{fig:Captions-comparision} illustrates that our GCS-M3VLT model provides accurate and contextually rich medical captions for retinal images and keywords. For various retinal conditions, including suprachoroidal hemorrhage, subretinal neovascularization (SRNV), and Goldmann-Favre syndrome, our model’s predictions closely match actual clinical descriptions, effectively capturing key diagnostic details and clinical context. Compared to existing methods \cite{shaik2024gated, huang2022non}, GCS-M3VLT consistently demonstrates superior accuracy in reflecting relevant retinal pathologies and patient details, highlighting its potential for enhancing clinical image interpretation and diagnosis.

\vspace{-5pt}
\section{Conclusion}

The proposed Guided Context Self-Attention based Multi-modal Medical Vision Language Transformer (GCS-M3VLT) advances retinal image captioning by integrating guided self-attention and multimodal fusion. This approach effectively captures contextual information from both retinal images and diagnostic keywords, enhancing the accuracy and coherence of clinical descriptions. Qualitative results demonstrate the model’s ability to localize lesion-specific features and adapt dynamically to various retinal abnormalities. The attention mechanisms improve focus on relevant regions, leading to more precise and relevant captions. Future work could explore aligning multi-modal embeddings in latent space while addressing missing keywords using zero-shot learning.

\printbibliography[title={References}]

\end{document}